\documentclass[lettersize,journal]{IEEEtran}
\usepackage{amsmath,amsfonts}
\usepackage{algorithmic}
\usepackage{array}
\usepackage[caption=false,font=normalsize,labelfont=sf,textfont=sf]{subfig}
\usepackage[caption=false,font=normalsize,labelfont=sf,textfont=sf]{subfig}
\usepackage{textcomp}
\usepackage{stfloats}
\usepackage{url}
\usepackage{verbatim}
\usepackage{graphicx}
\def\BibTeX{{\rm B\kern-.05em{\sc i\kern-.025em b}\kern-.08em
    T\kern-.1667em\lower.7ex\hbox{E}\kern-.125emX}}
\usepackage{balance}
\usepackage[ruled,vlined]{algorithm2e}
\usepackage{graphicx}
\graphicspath{{Figures/PDF/}{Figures/PNG/}}
\usepackage{subcaption}
\usepackage{booktabs}
\usepackage{siunitx}
\usepackage{texnames}
\usepackage{bm,bbm}
\usepackage{orcidlink}

\usepackage[backend=biber, style=numeric, citestyle=numeric, sorting=none, sortcites=true]{biblatex}
\usepackage{cleveref}
\usepackage{silence}
\usepackage{subcaption} 
\usepackage{float} 

\begin{document}
\title{Static and auto-regressive neural emulation of phytoplankton biomass dynamics from physical predictors in the global ocean}
\author{Mahima Lakra\IEEEauthorrefmark{1}\IEEEauthorrefmark{2}, Ronan Fablet\IEEEauthorrefmark{2}\IEEEauthorrefmark{4}, Lucas Drumetz\IEEEauthorrefmark{2}\IEEEauthorrefmark{4},  Etienne Pauthenet\IEEEauthorrefmark{3}, Elodie Martinez\IEEEauthorrefmark{3},

  \IEEEauthorblockA{\IEEEauthorrefmark{1,2}National Institute of Technology Karnataka, India}
  
  \IEEEauthorblockA{\IEEEauthorrefmark{2}IMT Atlantique, UMR CNRS LabSTICC,  Brest, France}

  \IEEEauthorblockA{\IEEEauthorrefmark{3}IRD, CNRS, University Brest, Ifremer, Laboratoire d’Océanographie Physique et Spatiale (LOPS), IUEM, Plouzané, France} 
    
  \IEEEauthorblockA{\IEEEauthorrefmark{4}Inria, Odyssey team,  Brest, France}

\thanks{}
    \thanks{}
}


\maketitle

\begin{abstract}
Phytoplankton is the basis of marine food webs, driving both ecological processes and global biogeochemical cycles.  Despite their ecological and climatic significance, accurately simulating phytoplankton dynamics remains a major challenge for biogeochemical numerical models due to limited parameterizations, sparse observational data, and the complexity of oceanic processes. Here, we explore how deep learning models can be used to address these limitations predicting the spatio-temporal distribution of phytoplankton biomass in the global ocean based on satellite observations and environmental conditions. First, we investigate several deep learning architectures. Among the tested  models, the UNet architecture stands out for its ability to reproduce the seasonal and interannual patterns of phytoplankton biomass more accurately than other models like CNNs, ConvLSTM, and 4CastNet. When using one to two months of environmental data as input, UNet performs better, although it tends to underestimate the amplitude of  low-frequency  changes in phytoplankton biomass. Thus, to improve predictions over time, an auto-regressive version of UNet was also tested, where the model uses its own previous predictions to forecast future conditions. This approach works well for short-term forecasts (up to five months), though its performance decreases for longer time scales. Overall, our study shows that combining ocean physical predictors with deep learning allows for  reconstruction and short-term prediction of phytoplankton dynamics. These models could become powerful tools for monitoring ocean health and supporting marine ecosystem management, especially in the context of climate change.

\end{abstract}


\section{Introduction}
\IEEEPARstart{P}{hytoplankton}, the microalgae inhabiting the ocean's upper layers, is crucial for fueling the marine food web and regulating carbon dioxide levels through photosynthesis. Assessing its variability in terms of biomass as well as its related drivers is therefore fundamental to identify emerging ecosystem dynamical issues, especially in the current context of global warming and climate changes. A vital prerequisite is to confidently assess the natural variability from seasonal, inter-annual to decadal time-scales in phytoplankton biomass, to then detect the anthropogenic signal. Despite advancements, satellite observation datasets are still too short to analyze decadal variability, and robust trend detection requires decades of continuous observations \cite{henson2010detection}. In addition, the uncertainties related to the complex balance of processes that control the phytoplankton fate, inducing processes imperfectly and diversely parameterized in bio-geochemical models, limit model’s ability to accurately simulate long-term variability of phytoplankton \cite{kwiatkowski2020twenty}. 

At global scale, phytoplankton variability has been widely associated with ocean and atmosphere dynamics (e.g., ; \cite{wilson2002global, kahru2010global, https://doi.org/10.1002/2015GB005216}). Thus, considering at first order that ocean and atmospheric dynamics drive phytoplankton biomass variability, over the last few years a range of machine learning techniques have been trained from physical predictors to reconstruct Chlorophyll-a concentration (Chl, used as a proxy of the phytoplankton biomass) time series. This approach was pioneered by \cite{19582008} through the application of linear canonical correlation analysis to sea surface temperature and sea surface height in the tropical Pacific Ocean (20°S-20°N). The Chl variability within 10° of the equator over $[1958-2008]$ was successfully reconstructed and decadal variations consistently with the Pacific Decadal Oscillation (PDO) were highlighted. The efficiency of data-driven approaches in reconstructing Chl variability from physical parameters at global scale was then demonstrated by investigating Support Vector Regression and Multi-Layer Perceptron techniques \cite{martinez2020reconstructing, rs14225669}. However, some regional bias in the reconstructions remained. Indeed, the relationships between phytoplankton and its physical surrounding environment were implicitly considered homogeneous in space, and training such models on a global scale didn’t allow one to consider known regional mechanisms. Other deep learning schemes, in particular Convolutional Neural Networks (CNNs), have shown a much greater ability to decompose and represent the spatial variations in Earth science reconstruction problems (e.g., ; \cite{sun2019combining, tc-14-1083-2020, PYO2021117483}), including studies focusing on Chl satellite observations  (e.g., ; \cite{yu2020global, roussillon2023multi, essd-15-5281-2023}).

Still, bias in capturing regional trends remains \cite{roussillon2023multi}. This may be due to 1) the lack of ad hoc predictors, 2) the fact that the machine or deep learning models used so far are not the most efficient. This latter point can be related with different model parameterizations and/or their inability to take into account the temporal dependencies of physical and bio-geochemical dynamics. For instance, among the physical processes that influence bloom development at mid and high latitudes, wintertime deep mixing plays a  key role in preconditioning the environment (i.e., allowing the near-surface replenishment of  nutrients) for a phytoplankton bloom the following spring \cite{gonzalez2018winter, williams2003physical}. 
Previous machine learning approaches for Chl reconstruction from physical predictors have been designed as a direct mapping from physics at time $t$ to Chl at time $t$, without considering time information explicitly. 

One of the objectives of this paper is to assess different ways of incorporating this information. We will mainly consider two settings: i) a "static" mapping first comparing four data driven methods considering no time dependencies, second with a window of several time samples around the current $t$ of the physical predictors to the Chl at time  $t$, ii) an "auto-regressive" mapping where, in addition, a past value of  Chl is explicitly considered in addition to the physical predictors. Every-time, the skills of the different methods and settings to capture the complex nonlinear relationship between physical predictors and satellite derived Chl are investigated.

\section{Data and Methods}  \label{sec2}
\subsection{Data}
We use  the Chl satellite derived  L3 product from the Ocean Colour-Climate Change Initiative OC-CCI v6 (hereafter referred to as Chl$_{Sat}$) which aggregates  observations from several satellite missions through a consistent preprocessing. This products improves the spatial and temporal coverage of Chl at global scale, when compared to a single mission. In addition, this dataset corrects inter-mission bias to product a climate quality long-term time-series. The following sensors are concerned: the Sea-viewing Wide Field-of-view Sensor (SeaWiFS, 1997-2010), the Moderate Resolution Imaging Spectroradiometer (MODIS, 2002-2019), the MEdium Resolution Imaging Spectrometer (MERIS, 2002-2012), and the Visible and Infrared Imaging Radiometer Suite (VIIRS, 2012-2019). Monthly observations over [1997-2017] on a 1/4° degree spatial grid have been downloaded from the Copernicus marine data store (https://doi.org/10.48670/moi-00282).
Eight physical predictors related to ocean and atmospheric dynamics allowing to provide light (surface  solar radiation-SSR) and nutrients to the upper layer, thus allowing phytoplankton to grow, are considered  (Table1).  Sea Level Anomaly (SLA) is a surface proxy of the thermocline variability. Sea Surface Temperature (SST) is related to surface stratification and the mixed layer depth (MLD) variability \cite{behrenfeld2006climate, martinez2009climate}. SST may also influence the phytoplankton physiology. Winds at 10m height (U10 and V10) are also related to mixing and MLD. Zonal (U) and meridional (V) surface currents may allow nutrient horizontal advection  \cite{messie2012global}. Finally, the Mean Dynamic Topography (MDT) represents the time-averaged sea surface height relative to the geoid, accounting for variations due to Earth's rotation and dynamic ocean circulation. MDT provides insights into large-scale ocean circulation patterns and nutrient availability at the surface of the ocean. 
The datasets for the eight physical predictors are resampled monthly from [1997-2017] on a $0.25^\circ \times 0.25^\circ$ grid between $50^\circ N$ and $50^\circ S$  to match the spatio-temporal resolution of  Chl$_{Sat}$ observations.  

\begin{table}[htbp]
\centering
\caption{Description of Physical Variables and Data Providers}
\label{Table1}
\resizebox{\columnwidth}{!}{%
\begin{tabular}{llll}
\toprule
\textbf{Variable} & \textbf{Description} & \textbf{Original Resolution} & \textbf{Data Provider} \\
\midrule
SLA  & Sea Level Anomaly & $0.25^\circ\times0.25^\circ$ daily & CNES-CLS \cite{pujol2023eu} \\
SST  & Sea Surface Temperature & $0.05^\circ\times0.05^\circ$ daily & Met Office \cite{good2020current} \\
SSR  & Surface Solar Radiation & $0.25^\circ\times0.25^\circ$ hourly & ERA5 reanalysis \cite{hersbach2020era5} \\
U    & Zonal surface current & $0.25^\circ\times0.25^\circ$ daily & OSCAR v2.0 \cite{dohan2021ocean} \\
V    & Meridional surface current & $0.25^\circ\times0.25^\circ$ daily & OSCAR v2.0 \cite{dohan2021ocean} \\
U10  & Zonal wind at 10 m & $0.25^\circ\times0.25^\circ$ hourly & ERA5 reanalysis \cite{hersbach2020era5} \\
V10  & Meridional wind at 10 m & $0.25^\circ\times0.25^\circ$ hourly & ERA5 reanalysis \cite{hersbach2020era5} \\
MDT  & Mean Dynamic Topography & $0.25^\circ\times0.25^\circ$ & CNES-CLS18 \cite{mulet2021new} \\
\bottomrule
\end{tabular}%
}
\end{table}

Some datasets, however, contain missing values (NaN), especially for oceanic variables over land areas. The NaN values in the input predictors are filled with zeros, ensuring the UNet model can process the data. This land mask is later removed from the loss function computation during the training, which ensure no effort is put on predicting the land mask. Given the log-normal distribution of Chl data, a logarithmic transformation is applied before incorporating it into machine learning algorithms. After reconstruction, the log(Chl) values are back-transformed to obtain Chl fields, which are then validated against Chl$_{Sat}$ observations. The term "log" denotes the natural logarithm. To enhance the stability of deep learning model training, each variable is standardized by subtracting its mean and dividing by its standard deviation, calculated over the period from January 1997 to December 2017. This normalization is a common practice in deep learning to improve training efficiency.
Following a standard deep learning experimental framework, we split the entire dataset  into three distinct, non-overlapping subsets for training, validation, and testing. The training dataset included years from 2002 to 2010, while the validation dataset used to monitor model generalization and parameter selection covered 20\% of the training dataset. Model performance is evaluated on the test dataset from 2012 to 2017. This configuration ensured sufficient temporal coverage to assess both seasonal and inter-annual variability which also includes the strong 2015/2016 El Niño event. Additionally, this split strategy minimizes the  autocorrelation between datasets and ensures consistency in the number of ocean color sensors used in the OC-CCI satellite dataset during the study period.

\subsection{Methods}
\begin{figure*}[htbp!]
    \centering
    \includegraphics[width = 18cm]{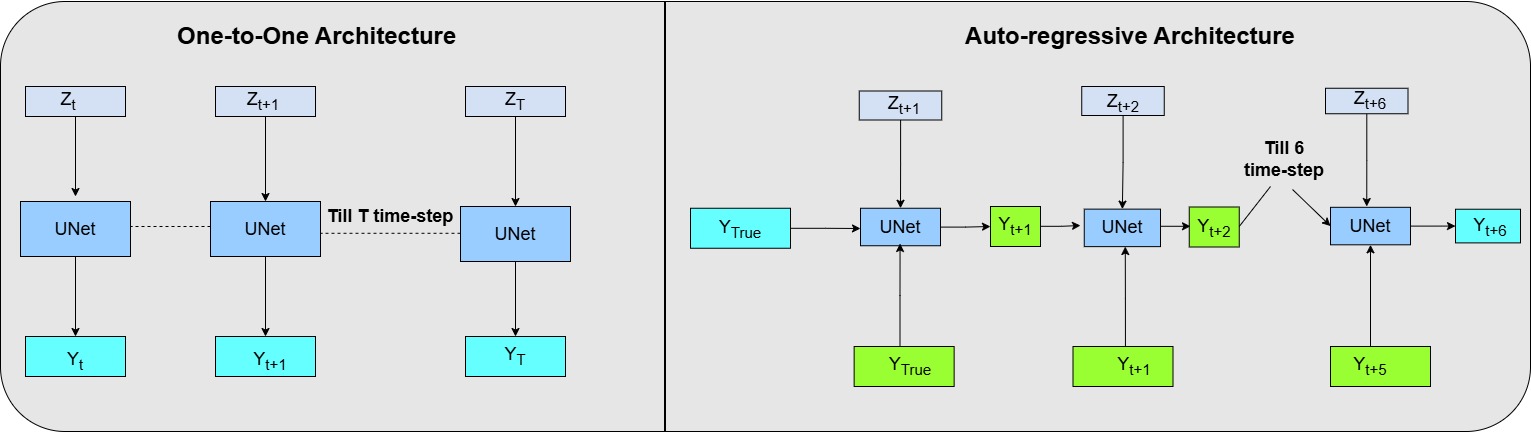}
    \caption{The UNet architecture is represented as an encoder-decoder architecture. The input state with size of $W \times L  \times 8 $, where dimension $8$ represents the number of physical predictors, the UNet generates outputs of size $W \times L \times 1$. Left: static (One-to-One architecture). Right: Auto-Regressive architecture. The weights of the UNet block are tied, i.e. the same parameters are used in every UNet block for each case.}
    
    \label{unet_model_mappings}
\end{figure*}

This section presents the considered neural architectures for the emulation of Chl dynamics from physical predictors. Let us denote by $Y_{[0,T]}=\left(Y(0),Y(1),\ldots,Y(T)\right )$ the targeted time series of Chl fields over a time window $[0,T]$, and by  $Z_{[0,T]}=\left(Z(0),Z(1),\ldots,Z(T)\right )$ the associated time series of multivariate fields comprising different physical drivers. 
$T$ corresponds to the number of monthly time windows, while $W \times L$ defines the spatial grid. At each space-time location, $Z$ encompasses eight physical predictors as described in the previous section. Overall, our goal is to predict time series $Y_{[0,T]}$ given $Z_{[0,T]}$. 
As sketched in Fig \ref{unet_model_mappings}, we may distinguish two main categories of neural emulators to address this issue. The first category proposed in \cite{roussillon2023multi,martinez_neural_2020} aims to learn a direct mapping between the physical predictors and the Chl states according to the following general equation:
\begin{equation}\label{eq: emulator 1}
    Y(t) = \mathbb{F}_\theta\left ( Z_{[t-\Delta^-,t+\Delta^+]} \right )
\end{equation}
where $\mathbb{F}_\theta$ is a neural network with trainable parameters $\theta$. $\Delta^-$ and $\Delta^+$ define the time window used for the physical predictors. This parameterization is illustrated in Fig \ref{unet_model_mappings} (left) with a UNet \cite{ronneberger2015u} for operator  $\mathbb{F}_\theta$ and $\Delta^-=\Delta^+=0$. Importantly, this first category of emulators does not rely on a sequential procedure to simulate a time series $Y_{[0,T]}$ given $Z_{[0,T]}$. Each time step $Y_{[0,T]}$ (or block of time steps if $\Delta^-$ or $\Delta^+\neq 0$) is predicted independently of the other time steps (or blocks).  

The second category of emulators relies on a Markovian parameterization inspired from the numerical schemes used to solve ocean biogeochemical models \cite{aumont2015pisces}. It predicts  the Chl state $Y(t)$ given the previous state $Y(t-\delta)$ and the physical predictors $Z(t-\delta)$ and $Y(t)$ according to
\begin{equation}\label{eq: emulator 2}
    Y(t) = \mathbb{G}_\theta\left ( Y(t-\delta),Z_{[t-\Delta^-,t+\Delta^+]} \right )
\end{equation}
where $\mathbb{G}_\theta$ is a neural network with trainable parameters $\theta$, and $\delta$ refers to the targeted time resolution of the time series $Y_{[0,T]}$. This auto-regressive parameterization is illustrated in Fig \ref{unet_model_mappings} (right) with a UNet for operator  $\mathbb{G}_\theta$. We may point out that operator $\mathbb{G}_\theta$ may rely on residual architectures as widely exploited in short-term neural forecasting approaches. Here, the time series $Y_{[0,T]}$ given $Z_{[0,T]}$ follows a sequential iterative procedure starting from an initial condition $Y(0)$.  

By definition, in a deterministic setting, the first category of emulators associated to parameterization \eqref{eq: emulator 1} can only account for the non-chaotic component of the dynamics of state $Y$ given physical forcings $Z$. By contrast, we expect the second category of emulators to perform better for short-term forecasts. However, beyond the Lyapounov time of the dynamics of $Y$, this improved performance is expected to vanish. This study aims to evaluate this behavior through numerical experiments as well as to benchmark different state-of-the-art neural architectures for operators $\mathbb{F}_\theta$ and $\mathbb{G}_\theta$, as detailed below. In the remainder, we refer to the first category of emulators as static emulators and the second category of emulators as auto-regressive emulators.

\paragraph{Baseline Convolutional Neural Network (CNN)}: 
Following \cite{roussillon2023multi}, we consider a baseline CNN architecture.  Based on formulation (\ref{eq: emulator 1}), the CNN used in \cite{roussillon2023multi} led to significant improvement compared with fully connected networks and kernel methods \cite{rs14225669}. Our CNN architecture comprises  2{\sc d} convolutional layers, each employing a $3 \times 3$ kernel size, a stride  $1 \times 1$, and  $1 \times 1$  padding. After each convolutional layer, a Rectified Linear Unit (ReLU) activation function is applied. 
 This architecture contains around 100,000 parameters. 

\paragraph{ Convolutional Long Short-Term Memory (ConvLSTM)}:
The ConvLSTM model as defined in \cite{shi2015convolutional} combines a LTSM architecture to capture short-long-term time dependendcies and convolutional layers to account for spatial patterns. Overall, it is among the state-of-the-art architecture to address  spatiotemporal datasets. The ConvLSTM architecture consists of $401 \times 1440 \times 3$ 3{\sc d} input tensors, where $3$ represents the length of the temporal sequence. The input sequence involves a temporal evolution over three time steps, resulting in a model output size of $401 \times 1440 \times 3$. The considered architecture stacks three ConvLSTM layers and the final layer maps the hidden state of the last ConvLSTM layer to the targeted output using a 2{\sc d} convolutional layer. 

\paragraph{Fourier ForeCasting Neural Network (4CastNet)}
Following  \cite{pathak2022fourcastnet},  we also explore the Adaptive Fourier Neural Operator (AFNO) model \cite{guibas2021adaptive}. This architecture is used for high-resolution input data and is among the state-of-the-art neural forecasting schemes. 
Specifically, it combines the Fourier Neural Operator (FNO) learning paradigm with a robust Vision Transformer (ViT). It was shown to be  effective in modeling complex partial differential equation (PDE) systems \cite{li2020fourier}. ViT has also demonstrated state-of-the-art performance in computer vision tasks.  The AFNO model overcomes the computational complexity of self-attention mechanisms in the ViT by reformulating spatial mixing as a global convolution in the Fourier domain, enabling efficient scaling for high-resolution data.  
We refer the reader to \cite{pathak2022fourcastnet} for a detailed presentation of the 4CastNet architecture.  Initially, the $401 \times 1440$ 2{\sc d} input tensors are partitioned into a set of $p \times p $ 2D patches  (e.g., $p=8$). Every patch is then represented as a d-dimensional token. The resulting sequence of tokens is augmented with positional encoding then it is subsequently processed by multiple AFNO layers.  Each AFNO layer processes an input tensor $Z$ by first applying a spatial mixing in the Fourier domain. This involves a 2D discrete Fourier transform, followed by a sparsity-constrained token weighting. The resulting representations in the Fourier domain are then inverse-transformed to the patch domain and combined with the original input via a residual connection. 

\paragraph{UNet}
UNets \cite{ronneberger2015u} are among the state-of-the neural architectures for image-to-image mapping tasks including among others image segmentation, image denoising, image super-resolution... \cite{yan2018shift, min2023d}.  As illustrated in Fig \ref{unet_model_mappings}, they exploit convolutional layers, pooling and up-sampling layers as well as skip connections. They provide an encoder-decoder architecture combining information extraction block at different scales.  In this study, we utilized a three-scale residual UNet architecture with bilinear blocks \cite{fablet2021learning} to account for the expected nonlinearities in ocean dynamics.

\paragraph{Diagnostics}
To assess the model’s capability to reproduce seasonal and non-seasonal variability, an Empirical Orthogonal Function (EOF) analysis is performed. The annual (monthly) mean Chl is first removed from the original time series to derive seasonal (non-seasonal) Chl anomalies, which were subsequently normalized by their standard deviations for both reconstructed Chl and Chl$_{Sat}$ . EOF is performed on Chl$_{Sat}$  and allow to retrieve the seasonal and non-seasonal spatial and temporal components. The reconstructed Chl anomaly time series is then projected onto the corresponding Chl seasonal and non-seasonal spatial patterns. The resulting temporal patterns, i.e., the principal components (PCs), are finally compared with those from Chl$_{Sat}$  using the Pearson correlation coefficient.

\section{Results} \label{sec4}
This section presents a comprehensive analysis of Chl static and auto-regressive emulators. In section \ref{sub1sec4}, we benchmark four static predictions. Subsequently, section \ref{impactofPP} investigates the preconditioning of the physical predictors, which is among the potential drivers of  phytoplankton spring blooms at high latitudes.  Section \ref{sub2sec4} delves into the auto-regressive approach using UNet based parameterizations. 

\subsection{Evaluation of static emulators}  \label{sub1sec4}
\begin{figure*}[htp]
\centering
  \includegraphics[width=15cm]{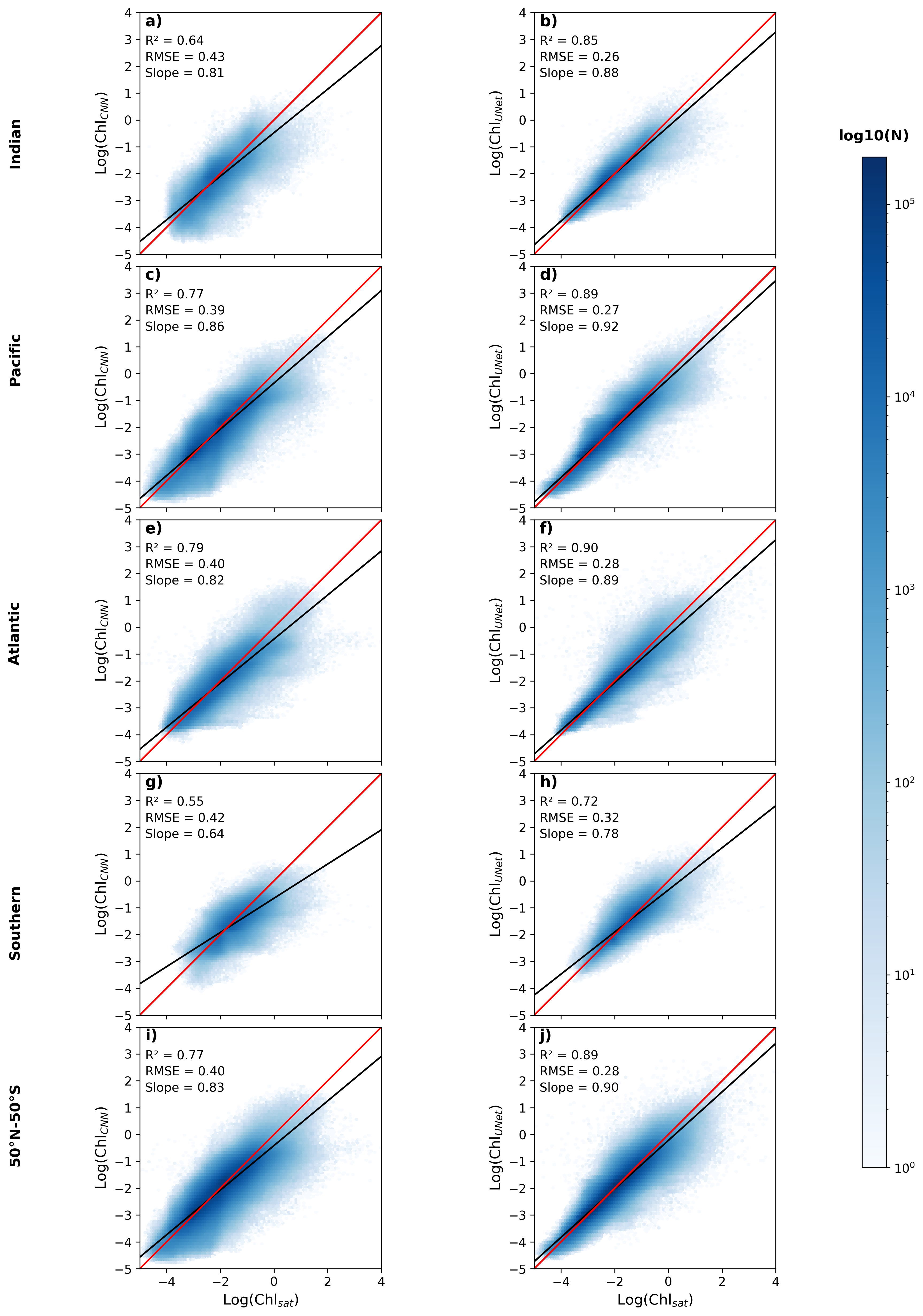}
  
    \caption{ Scatterplots of log(Chl$_{Sat}$)  vs.  reconstructed log(Chl) from the CNN (first column) and UNet (right column), across different oceanic basins over [2012-2017]   }
    \label{fig2}
\end{figure*}

Here, we assess the performance of static emulators defined by (\ref{eq: emulator 1}). We consider four distinct model architectures (CNN, ConvLSTM, 4CastNet, and UNet) for operator $F_\theta$ in (\ref{eq: emulator 1}) with $\Delta^-=\Delta^+=0$, {\em i.e.} using a single time-step for the physical predictors in (\ref{eq: emulator 1}). The predictive performance of these models are assessed comparing the reconstructed Chl against Chl$_{Sat}$ observations over the test time period $[2012-2017]$, independent from the training period $[2002-2010]$. 
Several diagnostics are produced to highlight each model's performances in capturing the spatial and temporal dynamics at different time scales. First, the quantitative evaluation are done calculating RMSE, the determination coefficient R$²$, the linear regression slope, and MAE over the period $[2012-2017]$ and averaged at global scale (50°N-50°S) ([Table \ref{table1}]). 
The convolutions-based CNN performs well in reconstructing satellite observations with R$²$= 0.77 and slope = 0.83, and RMSE=0.4 and MAE=0.30 between the reconstructed and Chl$_{Sat}$ values. However, the CNN is the less efficient of the four models with the weakest correlations and highest errors. Although these 4 models perform well in capturing spatio-temporal patterns, the CNN architecture has a smaller number of parameters compared to alternative models, which increase threefold computational time. The discrete nature of convolutional filters and pooling operations results in information loss and smoothing effects. Results improve from the CNN to the ConvLSTM model, then to the Fourier neural network-based 4CastNet, and finally to the UNet model which exhibits the strongest positive linear correlation (R$²$=0.88 and slope = 0.91) and the smallest error (RMSE = 0.28 and MAE = 0.20). These results highlight the UNet model’s ability to better learn the relationships from the training dataset.  
These metrics are also considered separately for the 4 oceanic basins (Indian, Pacific, Atlantic, and Southern Oceans) (Fig Supp \ref{fig_scatter_plot}). They highlight  the significant improvement in Chl reconstruction from the CNN to the ConvLSTM, 4CastNet and finally the UNet models which outperforms other models (see also Fig. \ref{fig2}).
\begin{table*}[htbp]
 \caption{The performance metrics for Chl static mapping across four distinct models (CNN, ConvLSTM, 4CastNet, and UNet)  assessed over the test period from January 2012 to December 2017 over 50°N--50°S. The best results are in bold.}
 \begin{center}
\begin{tabular}{p{2cm}|p{1 cm} p{.8cm} p{1cm}  p{1cm}|p{1 cm} p{.8cm} p{1cm}  p{2cm}|p{2cm} } 
 \hline \label{table1}

 Model& $R^2$ & RMSE & Slope & MAE& Corr. Seas. PC & Corr. Inter. PC & RMSE Seas.& RMSE Inter.& Computation Time\\
  \hline
 
 CNN  & 0.77 &0.40& 0.83& 0.30& 0.99& 0.83& 86.54 &124.62 & \bf{45 Min}\\
 \hline
 ConvLSTM&   0.84   &0.33& 0.86 &0.24& 0.99 & 0.78& 58.11& 134.49& 2h 15Min\\

 \hline
 4CastNet&   0.87   &0.30& 0.90 &  0.22 &  \bf{1} &0.87 &  \bf{39.96}& 136.9&2H\\
 \hline
 UNet&   \bf{0.88}   & \bf{0.28}&  \bf{0.90} & \bf{0.20}& 1&  \bf{0.89}& 58.39 &  \bf{117.80} &  1h 33Min\\

 \hline

\end{tabular}
\end{center}
\end{table*}

\begin{figure*}[ht!]
\centering

\subfloat{\includegraphics[width=0.45\textwidth]{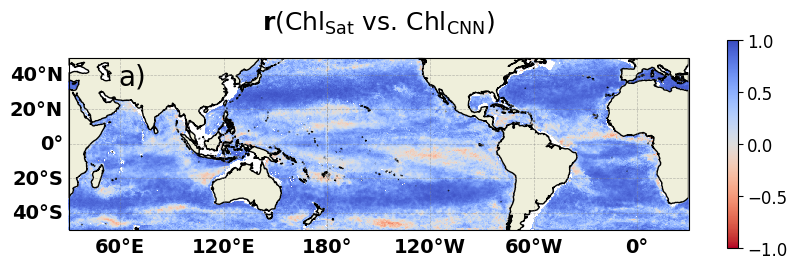}\label{fig3a}}%
\hfill
\subfloat{\includegraphics[width=0.45\textwidth]{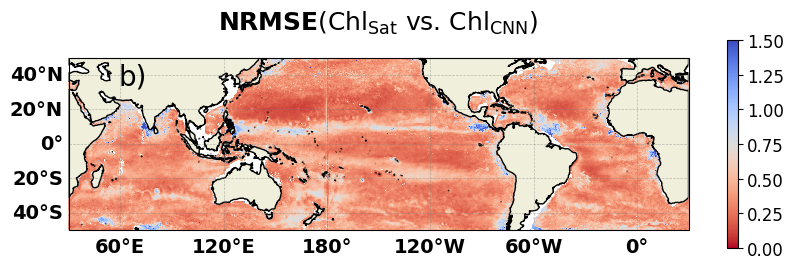}\label{fig3b}}


\subfloat{\includegraphics[width=0.45\textwidth]{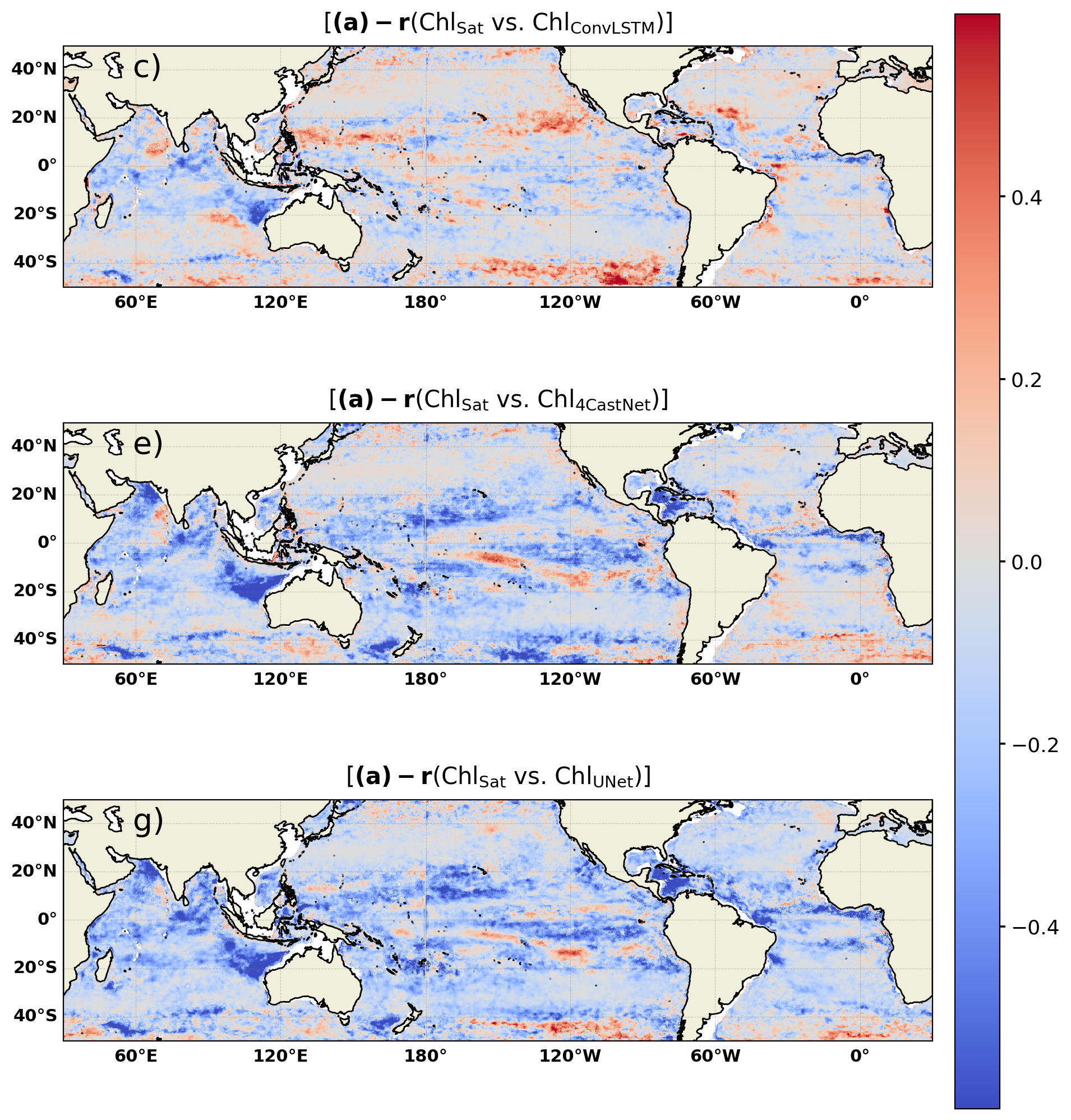}\label{fig3c}}
\hfill
\subfloat{\includegraphics[width=0.45\textwidth]{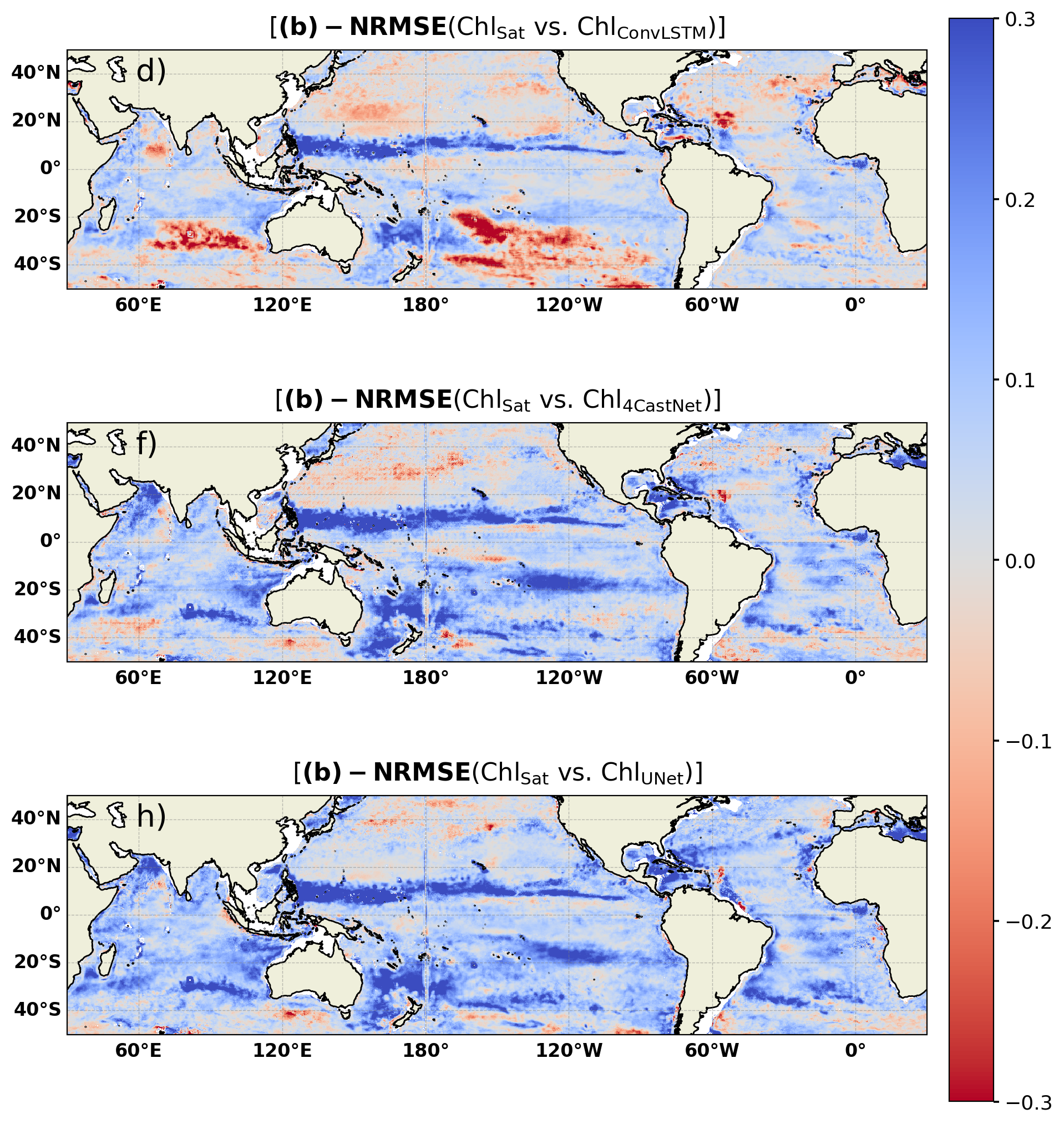}\label{fig3d}}

\caption{Correlation and NRMSE maps between Chl$_{Sat}$ vs. reconstructed Chl from CNN over the [2012--2017] testing time period. Correlation and NRMSE differences between ConvLSTM, 4CastNet, and UNet compared to CNN. For all the panels, the best results and improvements are in blue.}
\label{fig3}
\end{figure*}

Considering the temporal signal, it is more or less well reconstructed by the CNN as illustrated with correlation and NRMSE maps calculated over the test period between reconstructed Chl from CNN and satellite observations (Figs \ref{fig3}a and \ref{fig3}b). The CNN model succeeds in reconstructing Chl in the five subtropical gyres but it is less efficient in the tropical regions and at high latitudes where the ocean dynamics is more intense as in \cite{roussillon2023multi} .  Similar maps of correlations and NRMSE are performed between Chl$_{Sat}$ and reconstructed Chl from ConvLSTM, 4CastNet, and UNet. These maps are retrieved from those obtained with the CNN to highlight the improvements of these three models when compared to the former ones (in blue in Figs \ref{fig3}c to \ref{fig3}h). ConvLSTM improves the correlation in some regions, but worsens the NRMSE in the South Pacific reaching values up to  0.3 ( Figs \ref{fig3}c and \ref{fig3}d). In Figs \ref{fig3}e and \ref{fig3}f, the correlation and NRMSE difference between 4CastNet and CNN exhibit the improvement in 0°N and 20°S in the Indian Ocean.  Finally, the UNet improves correlations over most of the global ocean when compared to the CNN with  difference between 20°N and 20°S in the North Pacific region greater than 0.3 and reaching up to 0.6 (in blue in Fig  \ref{fig3}g and  Fig \ref{fig3}h). The correlation and NRMSE are here improved by using an encoder-decoder static emulator. The UNet architecture exhibits superior performance in predicting Chl$_{Sat}$  almost everywhere when compard to the CNN, except at high latitudes and on the equatorial Pacific along the High Nutrient Low Chlorophyll (HNLC) boundaries.

\begin{figure*}[ht!]
\centering

\begin{minipage}{1\textwidth}
  \centering
  \includegraphics[width=\textwidth]{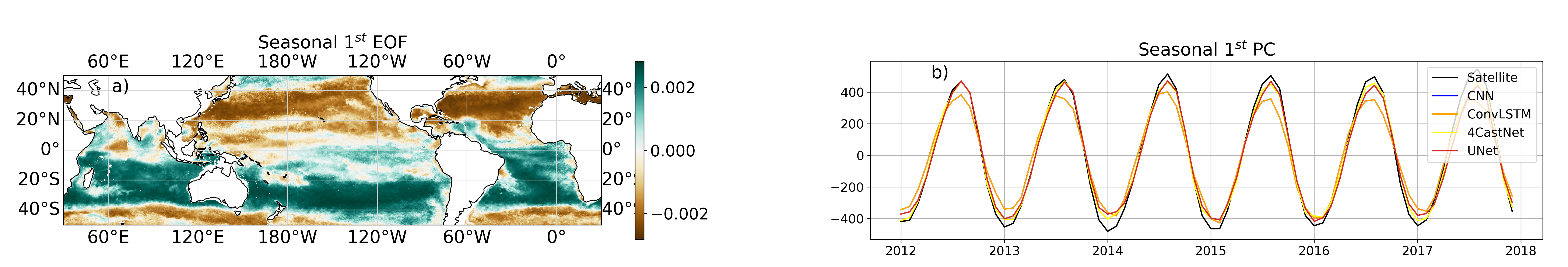}
  \label{fig4a}
\end{minipage}\hfill
\vspace{-0.3cm}
\begin{minipage}{1\textwidth}
  \centering
  \includegraphics[width=\textwidth]{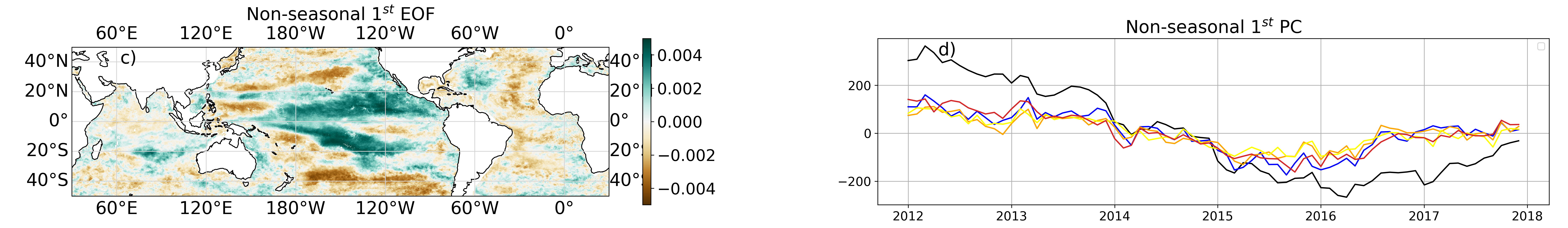}
 \label{fig4b}
\end{minipage}

\vspace{-0.3cm}

\begin{minipage}{1\textwidth}
  \centering
  \includegraphics[width=\textwidth]{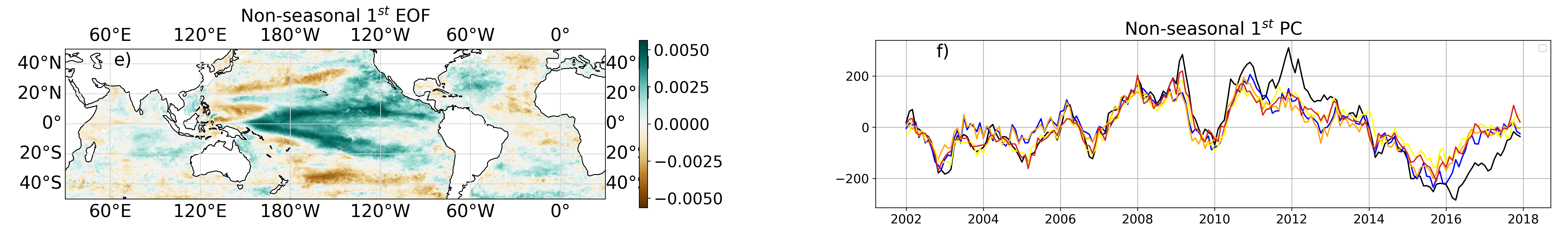}
  \label{fig4c}
\end{minipage}\hfill
\vspace{-0.3cm}
\begin{minipage}{1\textwidth}
  \centering
  \includegraphics[width=\textwidth]{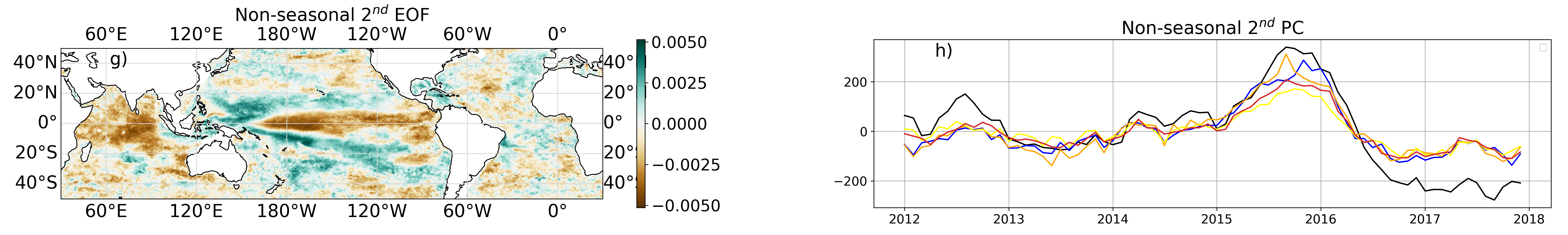}
  \label{fig4d}
\end{minipage}

\caption{EOF analysis based on seasonal and non-seasonal Chl$_{Sat}$ over the tested time period [2012--2017] and their projections on reconstructed Chl from the four static models. Panels (a),(c),(e),(g) show spatial patterns; (b),(d),(f),(h) show the corresponding principal components (PCs). EOF in (e),(f) is calculated on the non-seasonal signal over [2002--2018].}
\label{fig4}
\end{figure*}

The ability of the four models to capture the spatio-temporal variability of Chl$_{Sat}$ at seasonal and non-seasonal time-scales is also assessed performing EOF analysis over the test period 2012-2017.  The first mode of the Chl$_{Sat}$ based EOFs explain 29.9 \% and 8.6 \% of the total variance of the seasonal and non-seasonal signals, respectively. As expected, the seasonal cycle is out of phase between the two hemispheres (Fig \ref{fig4}a) and the corresponding PC (black line in Fig \ref{fig4}b) exhibits a sinusoidal pattern with a one-year period.  The four models succeed to reproduce the seasonal variability (Fig \ref{fig4}b), the best ones being the 4CastNet and UNet models with correlations up to 0.99 in both cases and a RMSE down to 39.9 and 58.3, respectively, between their projected PC and the Chl$_{Sat}$ PC (Table \ref{table1}). The first mode of the non-seasonal EOF shows a pattern related to what could be due to an inter-annual or lower frequency signal (Fig \ref{fig4}c and black line in Fig \ref{fig4}d). When considering the first mode over a longer time series (Fig \ref{fig4}e  and Fig \ref{fig4}f), this signal appears to be representative of a low-frequency variability. The second mode over the testing time period on its side, illustrates the inter-annual variability with the occurrence of the strong 2015-2016 El Nino event in the Pacific Ocean, which is pretty well reconstructed (Fig \ref{fig4}g and Fig \ref{fig4}h). While the UNet model shows a better ability to capture the Chl$_{Sat}$ seasonal an inter-annual signal than the CNN, ConvLSTM or 4CastNet (Fig \ref{fig4}f, Table \ref{table1}), all the models fail in reproducing the amplitude of this low frequency signal.

\subsection{Impact of time-lagged physical forcings}\label{impactofPP}

Here, we assess the performance of static emulators using different time-lagged configurations of the physical forcings. The considered time lag relates to the preconditioning of the environment on the Chl evolution. Based on the results in the previous section, we focus on UNet schemes.  
We consider four configurations as follows:
\begin{itemize}
\item $Y(t)=UNet[Z(t)]$ where Chl at time t only depends on physical predictors at the same time step.
\item $Y(t)=UNet[Z(t-1), Z(t)]$ where Chl depends on physical predictors at the same time step but also the step before (i.e., a short term preconditioning of the environment).
\item $Y(t)=UNet[Z(t-\Delta^-), \cdots, Z(t+\Delta^+)]$  where $\Delta^-=6$ and $\Delta^+=0$, which signifies that Chl depends from physical predictors over the six last time steps (i.e., as for instance the wintertime mixed layer depth in the North Atlantic may precondition the spring bloom 6 month after).
\item $Y(t)=UNet[Z(t-\Delta^-), \cdots, Z(t+\Delta^+)]$  where $\Delta^-=3$ and $\Delta^+=3$. Here, parameterization is inspired from numerical schemes used to solve ocean biogeochemical models.
\end{itemize} 

Considering the same diagnostics as in the previous section, the 2-month time window $\{Z(t-1), Z(t)\}$  is the best configuration (referred hereafter as UNet$_{Best}$) to reconstruct Chl  over the test period $[2012-2017]$. It is true at global scale (Table \ref{table_PC_Timewindow}) but also at basin-scale and regarding the spatial map of NRMSE and correlation difference when compared to the baseline Y(t)=UNet[Z(t)]  (Fig Supp S2). The EOF mode 1-based analysis which shows the projection of Chl$_{Sat}$ on reconstructed Chl illustrates that all the models reproduce the seasonal and non-seasonal variability (Table \ref{table_PC_Timewindow}) although here again the amplitude of the low frequency signal is underestimated (Fig not shown, but illustrated with a high RMSE in the non-seasonal PC). However, here again experimental results consistently indicate that a 2-month input predictor window Y(t)=UNet[Z(t-1), Z(t)] yields better performance when compared to other time window configurations in capturing the 1st mode of seasonal and non-seasonal variability amplitude.

\begin{table*}[htbp]
 \caption{Performance metrics for Chl static emulators obtained from physical predictors at varying time and evaluated over the test period [2012 - 2017] and (50°S-50°N). The best results are in bold.}
 \begin{center}
\begin{tabular}{p{3cm}|p{1 cm} p{.8cm} p{1cm}  p{1cm}|p{1.5 cm} p{1.5cm} p{1.5cm}  p{1.5cm} } 
 \hline \label{table_PC_Timewindow}

 Model& $R^2$ & RMSE & Slope & MAE& Corr. Seas. PC & Corr. Inter. PC & RMSE Seas. & RMSE Inter. \\
  \hline
 
 $Y(t)=UNet[Z(t)]$& 0.88 &0.28& 0.90& 0.20& 1& 0.89& 58.39 &117.80 \\
 $Y(t)=UNet[Z(t-1), Z(t)]$& \bf{0.89}   & \bf{0.27}& \bf{0.92} & \bf{0.19}& \bf{1}& \bf{0.91}& 40.25 &\bf{110.08} \\
 \hline
 UNet[Z(t-6),..., Z(t)]&   0.89   &0.27& 0.91 &0.20& 1 & 0.88& \bf{37.23}& 120.49\\

 \hline
 UNet[Z(t-3),..., Z(t)+3]&   0.88   &0.28& 0.87 &  0.20 &  {1} &0.86 &  {57.43}& 123.24\\

\end{tabular}
\end{center}
\end{table*}

\subsection{Benchmarking static and auto-regressive schemes}  \label{sub2sec4}
\begin{figure*}[h] 
    \centering
    \includegraphics[width=14cm]{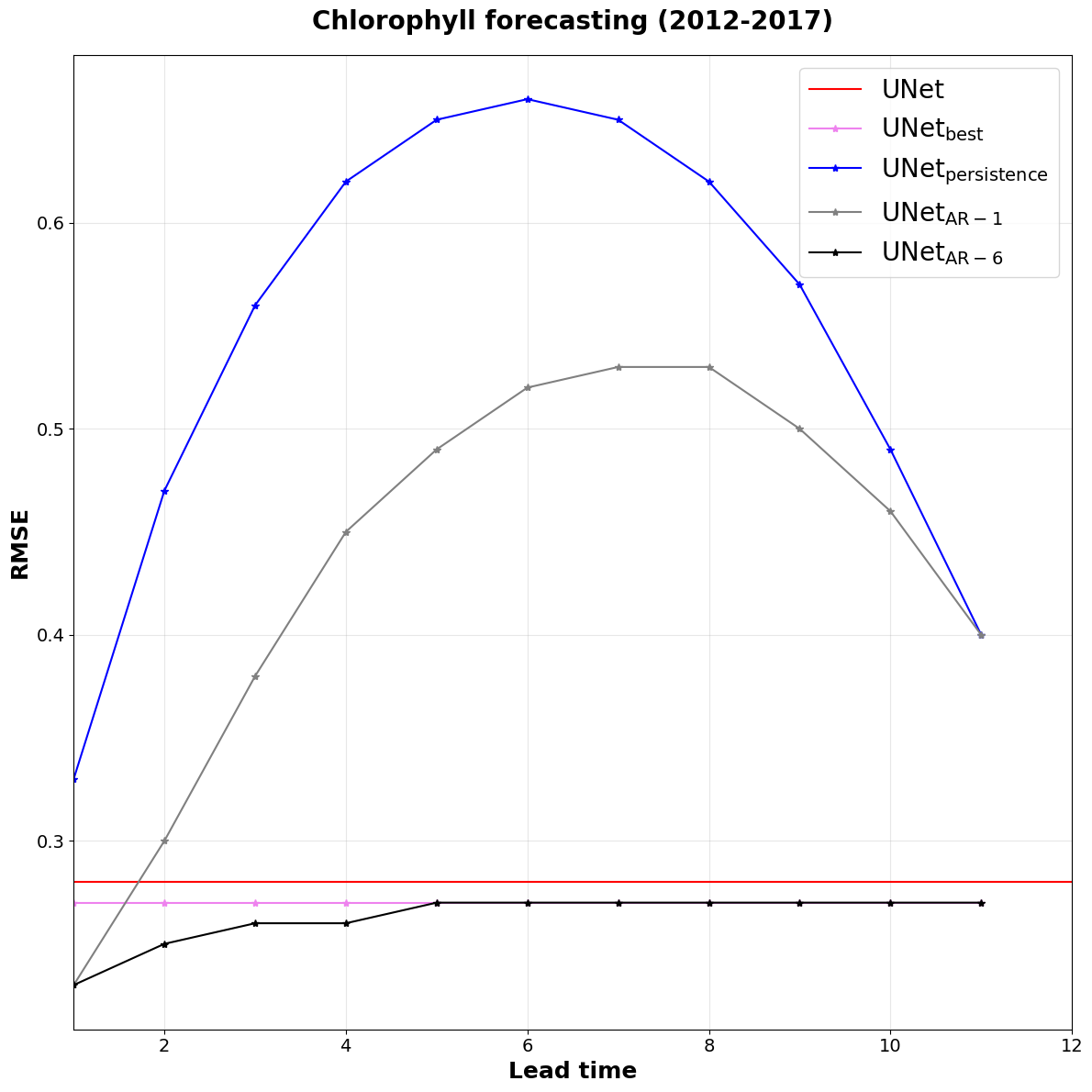} 
    \caption{ Trade-off between forecast accuracy and lead time for various parameterizations within an auto-regressive approach. It visually compares the RMSE achieved by UNet parameterization configurations at varying lead times (how far into the future the forecast predicts).} \label{forcast1}
    
    \label{unet}
\end{figure*}

Here, we aim to benchmark the performance of static  emulators defined by (\ref{eq: emulator 1}) and auto-regressive ones defined by (\ref{eq: emulator 2}). Following the results reported in Section \ref{impactofPP}, we compare the initial static UNet to UNet$_{Best}$ and to two auto-regressive UNet-based models defined according to (\ref{eq: emulator 2}) where operator $\mathbb{F}_\theta$ is parameterized by a UNet architecture similar to that of the best static baseline using $\{Z(t-1)), Z(t)\}$ as inputs, {\em i.e.} $\Delta^-=1$ and $\Delta^+=0$. The two auto-regressive schemes differ in the number of iterative roll-outs used during the training phase: UNet$_{AR-1}$ is trained using just one forecasting step, whereas UNet$_{AR-6}$ is trained to minimize the forecasting over a 6-month forecasting window. For benchmarking purposes, we also include in our experiments a persistence baseline which predicts the Chl state at time (t +1) as the known state at time (t), hereafter referred to as UNet$_{Persistence}$  We synthesize these experiments in Tab.\ref{table2} and Fig \ref{forcast1}. 

\begin{figure*}[ht!]
\centering

\begin{minipage}{0.5\textwidth}
    \centering
    \includegraphics[width=\textwidth]{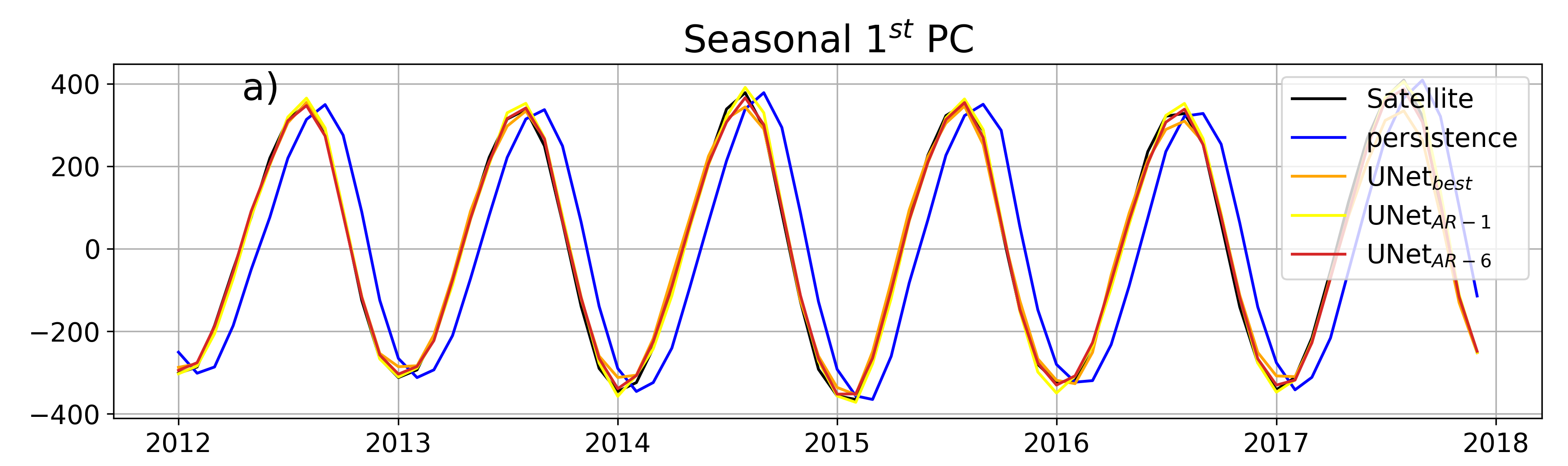}
    \label{fig6:s4fig1}
\end{minipage}\hfill
\begin{minipage}{0.5\textwidth}
    \centering
    \includegraphics[width=\textwidth]{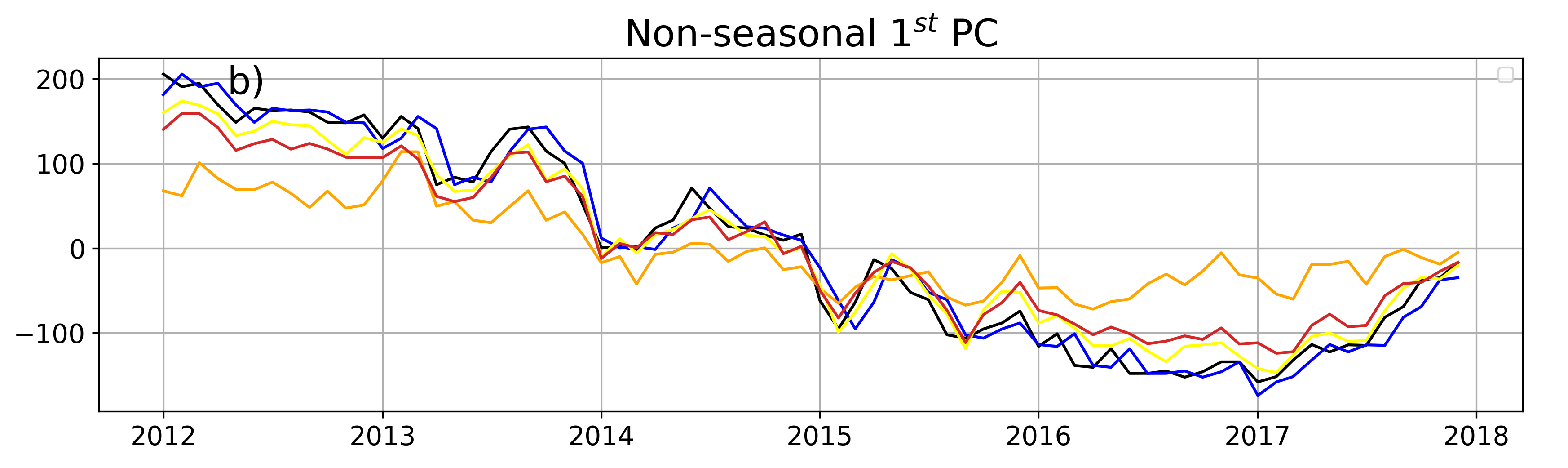}
    \label{fig6:s4fig2}
\end{minipage}

\vspace{-0.25cm}

\begin{minipage}{0.5\textwidth}
    \centering
    \includegraphics[width=\textwidth]{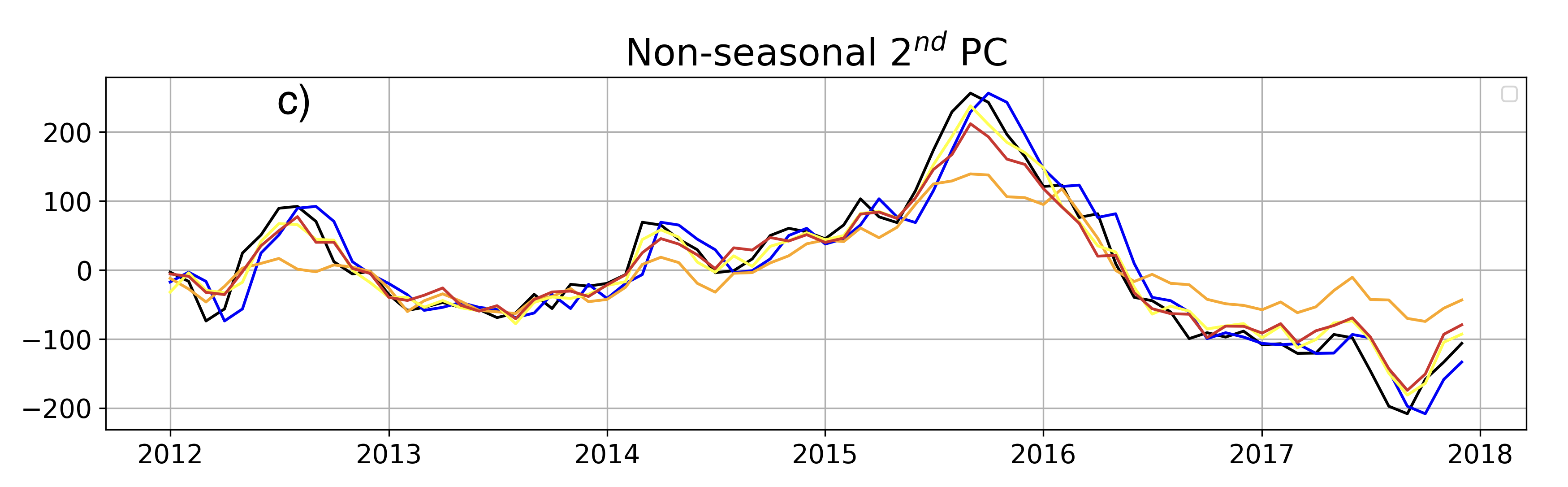}
    \label{fig6:s4fig4}
\end{minipage}

\caption{EOF analysis based on seasonal and non-seasonal Chl$_{Sat}$ over the test time period [2012--2017] and their projections on reconstructed Chl from the four static models. The corresponding spatial patterns are the same than in Fig. 4, b, d, h. }
\label{fig6}
\end{figure*}

The greater forecasting performance of the auto-regressive neural schemes compared with a persistence baseline as well as the impact of a 6-month roll-out during the training phase are clearly highlighted. The one-month roll-out training setting poorly generalizes when forecasting for two months or more. Whereas its forecasting performance improves that of the static UNet$_{Best}$ scheme for a 1-month lead-time, the performance rapidly worsens and becomes lower than that of the reference from a 2-month lead-time. 
As expected, the 6-month roll-out  auto-regressive scheme ({\em i.e.}, UNet$_{AR-6}$) leads to the best performance and outperforms the UNet$_{Persistence}$ for all lead times by 50\% or more in terms of RMSE. These results are confirmed spatially at global scales (Fig Supp3). It also outperforms the UNet$_{Best}$ scheme for the first lead-times up to 4-month with a decaying relative gain as the forecasting lead-time increases. From a 5-month lead-time, we do not report any significant difference between the best auto-regressive scheme and the static one. This pattern relates to the phytoplankton chaotic dynamics. On average and on a global scale, our results suggest a Lyapounov time of a few months for Chl dynamics. This seems in line with the characteristic time scales of Chl processes, such as spring blooms in the mid-latitudes. Considering the ability of the auto-regressive approaches to reconstruct seasonal and non-seasonal Chl variability, still at lead time 1, the seasonal cycle in PC1 (Table \ref{table2}) is almost perfectly captured by the two regressive approaches with a correlation with the satellite PC of 1. The most striking results is their performance result in their ability to capture the amplitude of the low frequency signal (Fig  \ref{fig6}), especially for the UNet$_{AR-1}$ (RMSE between non-seasonal reconstructed Chl from UNet$_{AR-1}$ and satellite = 33,9, vs. 55,8 from the UNet$_{AR-6}$ and 110 for UNet$_{Best}$).

To assess the long-term forecasting capabilities of the proposed auto-regressive approach, we extended the forecast lead times beyond six months till 11 months. The experimental results contrast the performance of the optimal one-to-one mapping approach against the best short-term (i.e. from 1 to 6 months) parametrization. As shown in Tables 2 and 4, the UNet$_{Best}$ model and the UNet$_{AR-6}$ parameterization configuration exhibit superior performance regarding basin-wide metrics. Here again, to further quantify the model's proficiency at extended time scales, we examined its ability to reconstruct satellite Chl (scatter plots at global and basin scales), capture spatial patterns (NRMSE and correlation maps) and Chl temporal dynamics (EOFs) (Figs not shown as results are illustrated in Table \ref{table2}). A comprehensive comparison was conducted between UNet$_{Best}$ model and the UNet$_{AR-6}$ parameterization, both evaluated against Chl$_{Sat}$ data at forecast lead times exceeding six months. Although less efficient at higher lead time than 1 or 6 months, all the diagnostics mentioned above (except for the EOF) reveal a key finding: at longer lead times than 6 months, there is no observable difference between the UNet$_{Best}$ model and the UNet$_{AR-6}$ auto-regressive approach. Specifically, the UNet$_{AR-6}$ consistently maintains the metrics considered, regardless of whether the forecast is 6 or 11 months ahead. The UNet$_{AR-6}$ configuration demonstrates robust performance in forecasting future scenarios, particularly at longer lead times. Considering the EOF analysis, the UNet$_{AR-6}$ at lead times of 6 to 11 months effectively captures the seasonal variability but exhibits limitations in accurately representing the amplitude of the low frequency signal in the non-seasonal mode 1. 

\begin{table*}[h!]
 \caption{ Performance metrics for Chl static and auto-regressive approaches for lead time 1 to 12  over the testing period [2012-2017] and (50°S-50°N). The best resulats are in bold.}
 \begin{center}

\begin{tabular}{p{3.5 cm} p{2.5cm}|p{.5 cm} p{.5cm} p{.5cm}  p{.5cm} p{.5cm} p{.5cm} p{.5cm} p{.5cm} p{.5cm} p{.5cm} p{.5cm} p{.5cm}} 
 \hline \label{table2} 
 UNet Model& &  & Lead &  Time& & & & & & & & &\\
 \hline
Different Parametrization & Metrics &  1& 2 &3 &4&5&6&7&8&9&10&11\\
\hline

UNet$_{best}$  &  $R^2$ & \bf{0.89} &   \\
 &  RMSE & \bf{0.27} &  \\
 &  Slope & \bf{0.92} &  \\
 &  MAE & \bf{0.19} &     \\
  &  Corr. Seas. PC &  \bf{1}& \\
 &  Corr. Inter. PC & \bf{0.91} &  \\
 &  RMSE Seas. & \bf{40.25} &   \\
 &  RMSE Inter.  & \bf{110.08} &   \\
 \hline
 UNet$_{Persistence}$    &  $R^2$ & \bf{0.84} & 0.69 & 0.58 & 0.50 & 0.46 & 0.44 & 0.46 & 0.49 & 0.56 & 0.66 & 0.74  \\
&  RMSE & \bf{0.33} & 0.47 & 0.56 & 0.62 &0.65 & 0.66 & 0.65 & 0.62 & 0.57 & 0.49 & 0.44 \\
 &  Slope & \bf{0.92}  & 0.83 &0.76   & 0.71 & 0.68 & 0.66 & 0.67 & 0.70 & 0.75 & 0.81 & 0.87    \\
 &  MAE & \bf{0.24} & 0.35  & 0.43 & 0.48 & 0.50 & 0.51 & 0.51 & 0.48 & 0.44 & 0.37 & 0.29 \\
 &  Corr. Seas. PC & \bf{0.86} & 0.49  & 0.01 & $-0.49$& $-0.85$ &$-.98$ &$-0.85$ & $-0.49$& 0.01& 0.498& 0.86\\
 &  Corr. Inter. PC & \bf{0.98} & 0.95  & 0.93 &0.90 & 0.90 &0.91 & 0.88&0.86 & 0.83&0.81 & 0.81 \\
 &  RMSE Seas. & 130.6 &  248.8 &350.1  & 427.0& 474.7& 491.8&470.7 & 421.7&344.3 & 244.5& \bf{128.61}\\
 &  RMSE Inter.  &\bf{21.3}  &  33.66 & 41.12 &46.1 & 47.89&47.9 & 52.8& 57.11&62.42 & 67.02& 68.67\\
 \hline

 UNet$_{AR-1}$ &  $R^2$ & \bf{0.92} & 0.87 & 0.78 & 70& 0.64& 0.60& 0.59 & 060 & 0.64 & 0.70 & 0.78\\
 &  RMSE &  \bf{0.22}& 0.30 & 0.38 & 0.45 & 0.50 & 0.53 & 0.54 & 0.54& 51 & 0.46 & 0.39 \\
 &  Slope &   \bf{0.95} & 0.88 & 0.86 & 0.81 & 0.77 & 0.75 & 0.74 & 0.76 & 0.79 & 0.84 & 0.89\\
 &  MAE & \bf{0.16} & 0.22 & 0.29 & 0.34 & 0.38 & 0.41 & 0.41 & 0.41& 0.41 &  0.34 & 0.29\\
 &  Corr. Seas. PC & \bf{1} & 1  &  0.60 & $-.14$ & $-.88$ & $-.78$ & $-.45$& $-.09$& 0.26 & 0.56 & 0.79\\
 &  Corr. Inter. PC & \bf{0.99} & 0.87  & 0.98  & 0.96& 0.95& 0.94& 0.93 & 0.92 & 0.91 & 0.89 & 0.88\\
 &  RMSE Seas. &\bf{18.85}  & 136.9 & 274.8 & 383.5& 467.1& 518.5 & 533.4 & 512.1 & 455.7& 368.5& 257.6\\
 &  RMSE Inter.  & \bf{33.89} &   39.95 & 41.75 & 51.20& 58.33& 61.50& 64.85& 69.85& 77.13 & 83.19 & 89.03\\

 \hline
UNet$_{AR-6}$&  $R^2$ & \bf{0.92} & 0.90 & 0.90 & 0.90 &0.89 &  0.89 & 0.89 &0.89 & 0.89 & 0.89 &0.89 \\
 &  RMSE & \bf{0.23}  &  0.25&  0.25 & 0.26 & 0.27 & 0.27 & 0.27 & 0.27 & 0.27 & 0.27 & 0.27\\
 &  Slope & \bf{0.94} & 0.93  & 0.93 & 0.94 & 0.94 & 0.94 & 0.94 & 0.94 & 0.94 & 0.94 & 0.94 \\
  &  MAE & \bf{0.16} & 0.18 & 0.18 & 0.19 & 0.19 & 0.19 & 0.19 & 0.19 & 0.19 & 0.19 & 0.19  \\
  &  Corr. Seas. PC & \bf{1} & 1  & 1  & 1& 1& 1&1 &1 &1 &1 & 1 \\
 &  Corr. Inter. PC &  \bf{0.99}& 0.98  & 0.96 &0.94 & 0.93 & 0.92& 0.91&0.91 &0.90 &0.89 & 0.89  \\
 &  RMSE Seas. & \bf{18.43} & 24.22  & 27.96  & 29.58& 30.65& 30.25& 29.16&29.53 &30.9 & 30.9 & 30.77\\
 &  RMSE Inter.  &\bf{55.79} &  77.59 & 87.95 & 95.08 & 98.70 & 101.6&104.6 & 105.6&106.6 & 107.4 & 108.19 \\

 \hline
 \end{tabular}
\end{center}
\end{table*}

\section{Conclusion} \label{con} 
A detailed study of deep learning-based architectures has been discussed for mapping Chl$_{Sat}$  from oceanic and atmospheric physical predictors in the global ocean.  Integration of physical forcing and encoder-decoder architecture, UNet adeptly navigates the complex spatio-temporal phytoplankton dynamics  better than the CNN, ConvLSTM, and  4CastNet.  The UNet well captures the Chl spatial and seasonal and interannual variability when considering the one-month physical predictor time window. However, the amplitude of the EOF non-seasonal 1st mode, likely representing the low frequency variability, is underestimated. The 2-month input predictor time window compared to other time window is the best when considering static UNet configurations, although still not resolving the low frequency amplitude bias. These findings  highlight the inherent trade-offs between temporal and spatial accuracy in Chl$_{Sat}$ reconstruction. The addition of the auto-regressive approach results in two emulators associated with a 1- and 6-months roll-out during the training phase. The later emulator is able to make  Chl predictions up to six months, and also better reproduce the amplitude of the low-frequency signal. However beyond six months, this improved performance vanished. 
By leveraging large, heterogeneous datasets—including satellite remote sensing and physical model outputs— deep learning can detect nonlinear patterns and relationships that traditional numerical models may overlook. Our comprehensive assessment  demonstrates that both static and auto-regressive approaches with UNet settings are capable to retrieve spatio-temporal patterns of Chl$_{Sat}$.  Our results underscore the potential of neural emulators with physical environmental conditions for projecting long term satellite-observed Chl measurements. Furthermore, they highlight how deep learning methods can be used to fill observational gaps, enhance spatial and temporal resolution, and provide more accurate reconstructions of phytoplankton variability. This approach also offers valuable insights for marine ecology and sustainable management practices in response to climate change.

\section{Data availability statement}
The datasets presented in this study can be found in GitHub repositories. The details of the repository/repositories are as follows: \url{https://github.com/ANR-DREAM/auto-regressive_paper}

\section*{Acknowledgments} 
This research was funded by the Agence Nationale de la Recherche (ANR) under the project ANR‐22‐CE56‐0002‐01
We would like to thank Joana Roussillon for providing his Python program for CNN initial research.  We acknowledge the project team of Ocean Colour Climate Change Initiative for generating and sharing the merged datasets on Chlorophyll-a concentrations. We benefited from HPC and GPU resources from CPER AIDA GPU cluster supported by The Regional Council of Brittany and FEDER.

\section{Conflict of interest}
The authors declare that the research was conducted in the absence of any commercial or financial relationships that could be construed as a potential conflict of interest.

\clearpage

\section{Supplementary}
\setcounter{figure}{0}
\renewcommand{\thefigure}{S\arabic{figure}}
\begin{figure*}[ht]
\centering
  \includegraphics[width=15cm]{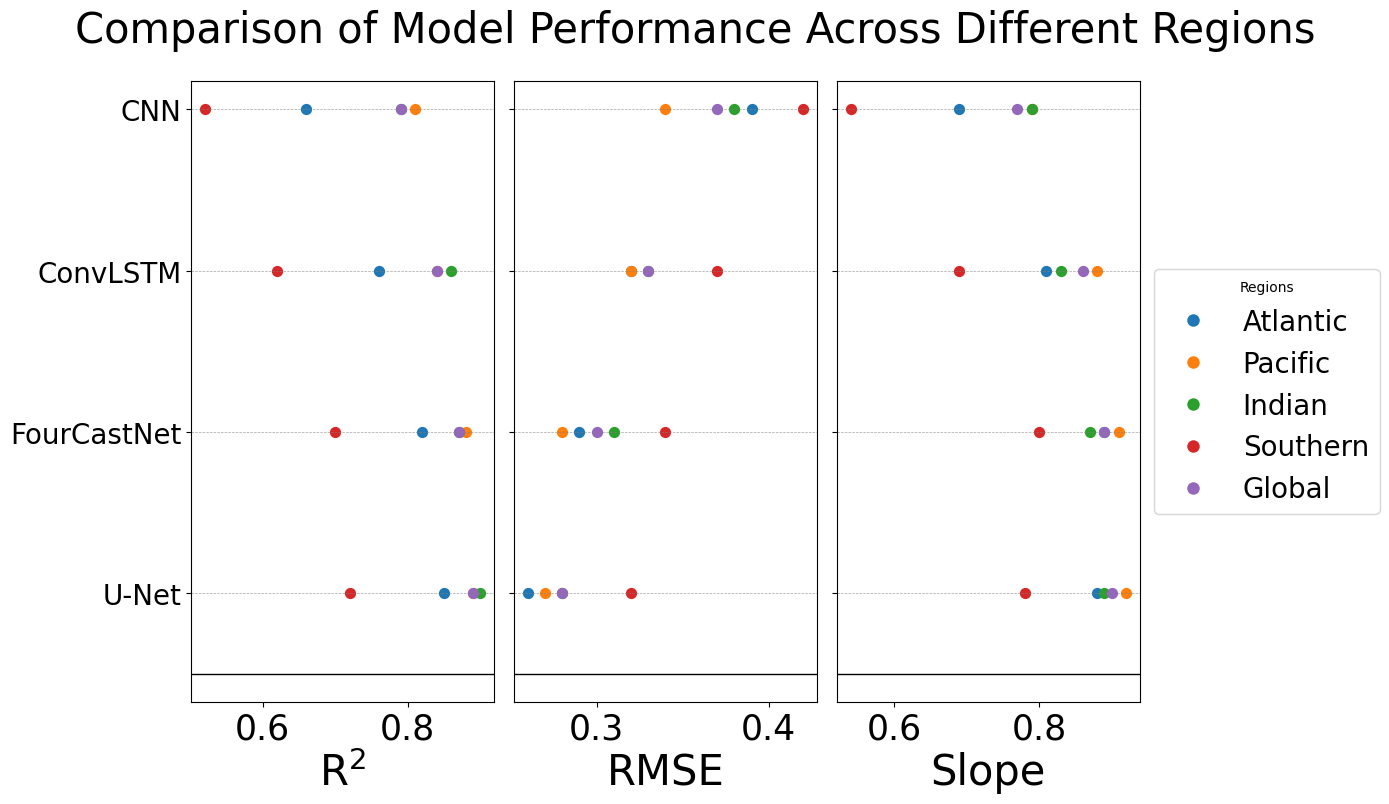}
  
    \caption{ The scatterplots illustrate the comparison between the satellite log (Chl$_{Sat}$)  and the reconstructed log(Chl) from CNN, ConvLSTM, 4CastNet, and UNet methods over the testing time period [2012-2017].}
    \label{fig_scatter_plot}
\end{figure*}

\begin{figure*}[ht!]
\centering

\subfloat{%
  \includegraphics[width=0.45\textwidth]{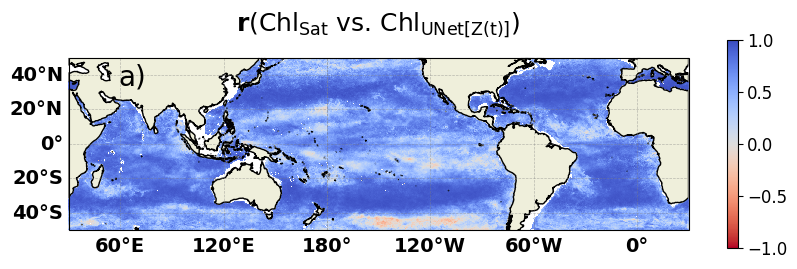}%
  \label{fig:s2fig1}%
}\hfill
\subfloat{%
  \includegraphics[width=0.45\textwidth]{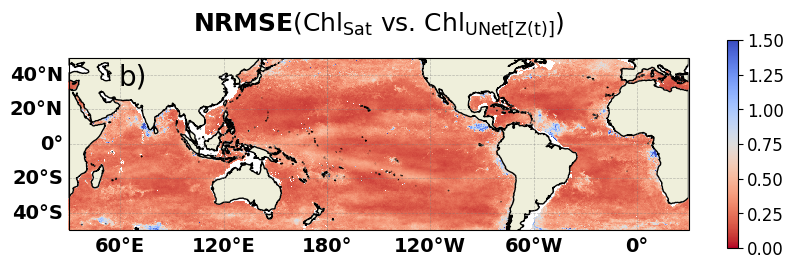}%
  \label{fig:s2fig2}%
}


\subfloat{%
  \includegraphics[width=0.45\textwidth]{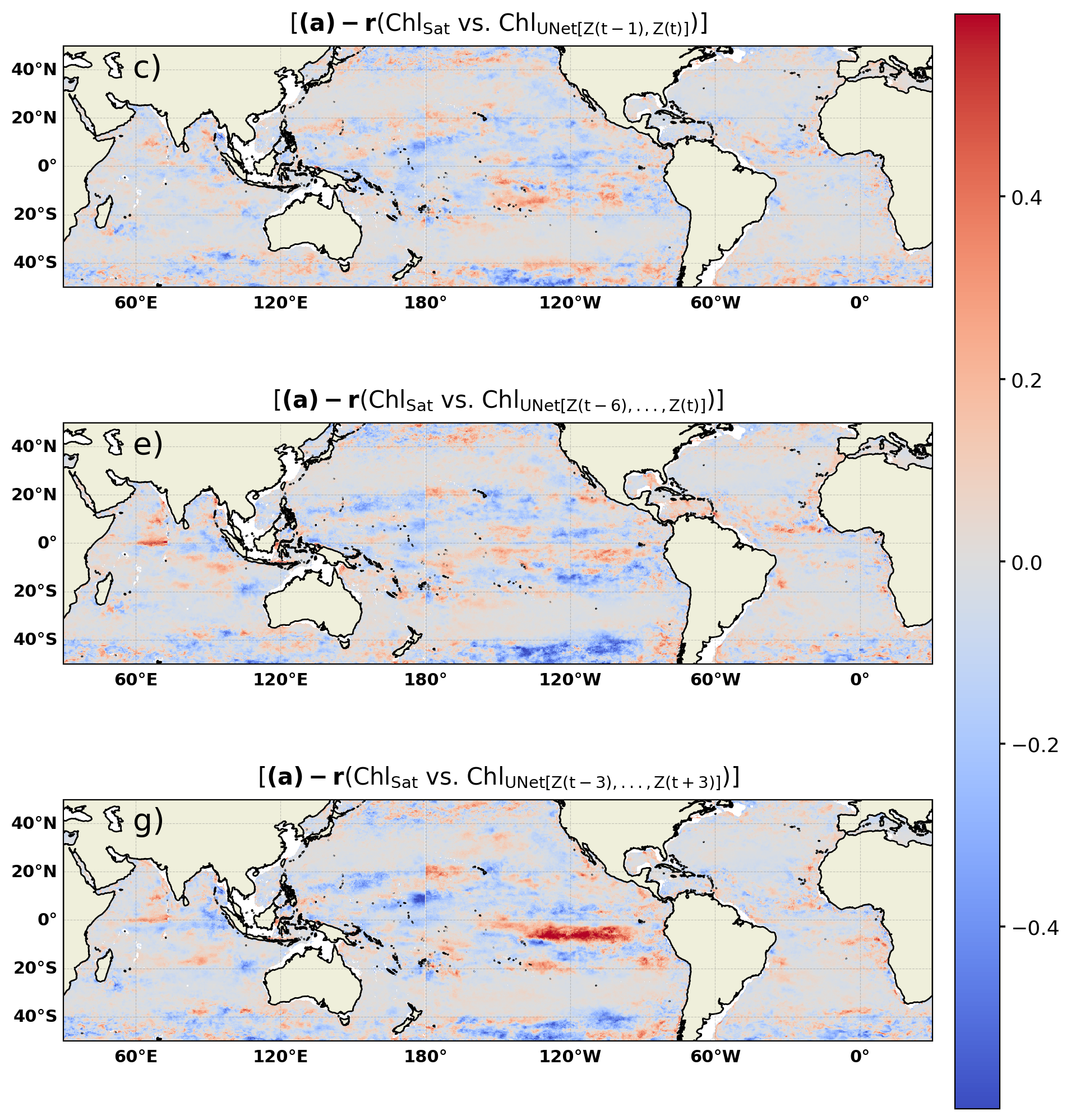}%
  \label{fig:s2fig3}%
}\hfill
\subfloat{%
  \includegraphics[width=0.45\textwidth]{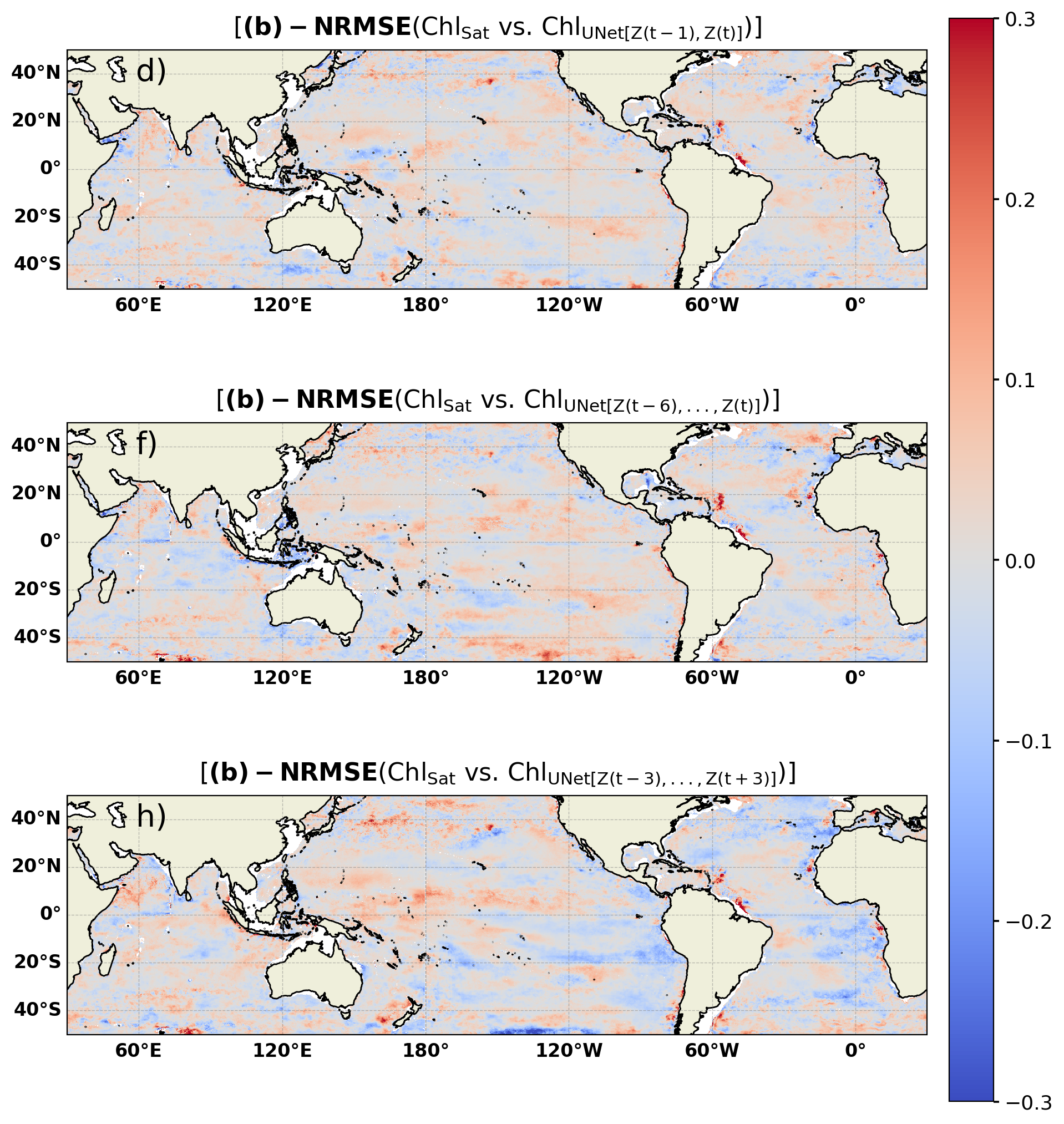}%
  \label{fig:s2fig5}%
}

\caption{(a) Correlation and (b) NRMSE maps between Chl$_{Sat}$ vs. reconstructed Chl from Chl$_{UNet[Z(t)]}$ over the [2012--2017] testing time period. (c) Correlation and (d) NRMSE differences between Chl$_{UNet[Z(t)]}$ and Chl$_{UNet[Z(t-1), Z(t)]}$. (e) Correlation and (f) NRMSE differences between Chl$_{UNet[Z(t)]}$ and Chl$_{UNet[Z(t-6), ..., Z(t)]}$. (g) Correlation and (h) NRMSE differences between Chl$_{UNet[Z(t)]}$ and Chl$_{UNet[Z(t-3), ..., Z(t+3)]}$. For all the panels, the best results and improvements are in blue.}
\label{fig12_lead1}
\end{figure*}

\begin{figure*}[ht!]
\centering

\subfloat{%
  \includegraphics[width=0.45\textwidth]{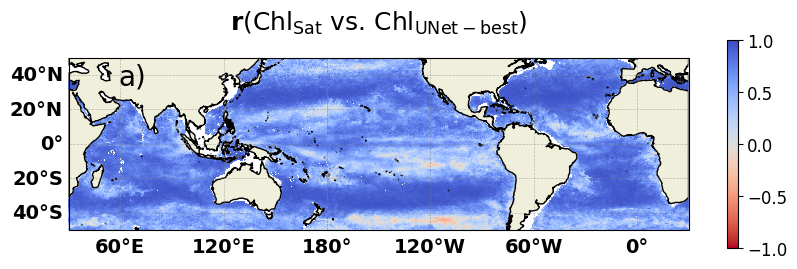}%
  \label{fig:s3fig1}%
}\hfill
\subfloat{%
  \includegraphics[width=0.45\textwidth]{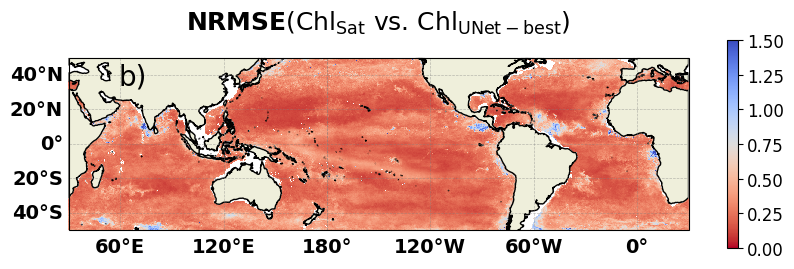}%
  \label{fig:s3fig2}%
}


\subfloat{%
  \includegraphics[width=0.45\textwidth]{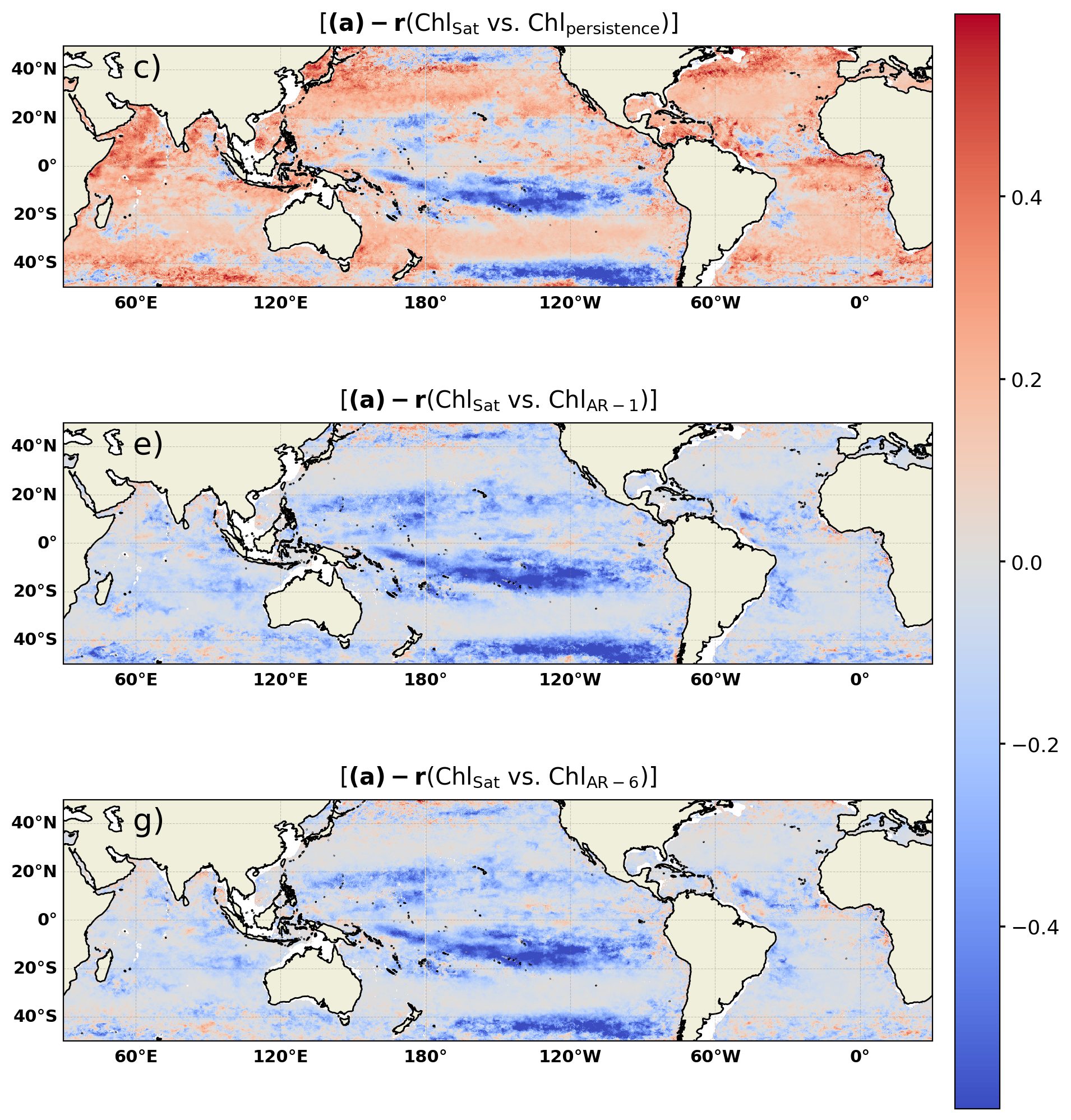}%
  \label{fig:s3fig3}%
}\hfill
\subfloat{%
  \includegraphics[width=0.45\textwidth]{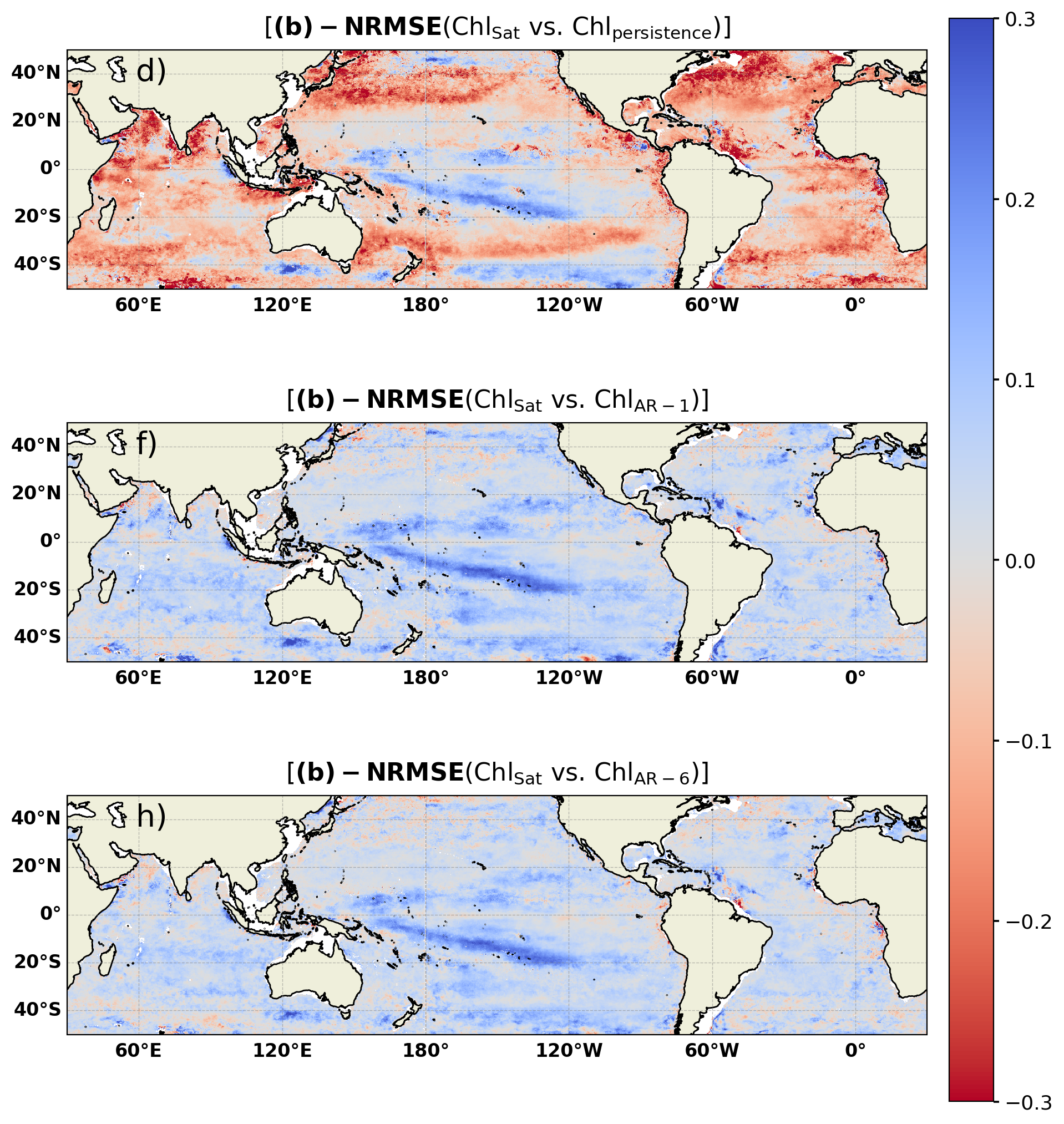}%
  \label{fig:s3fig5}%
}

\caption{(a) Correlation and (b) NRMSE maps between Chl$_{Sat}$ vs. reconstructed Chl from Chl$_{UNet-Best}$ over the [2012--2017] testing time period. (c) Correlation and (d) NRMSE differences between Chl$_{UNet-Best}$ and Chl$_{Persistence}$. (e) Correlation and (f) NRMSE differences between Chl$_{UNet-Best}$ and Chl$_{AR-1}$. (g) Correlation and (h) NRMSE differences between Chl$_{UNet-Best}$ and Chl$_{AR-6}$. For all the panels, the best results and improvements are in blue.}
\label{fig13_lead1}
\end{figure*}


\printbibliography

\begin{IEEEbiographynophoto}{Mahima}
Dr.\ Mahima is currently an Assistant Professor at the National Institute of Technology Karnataka (NITK), Surathkal, India. She specializes in computer vision and deep learning applications in oceanography. She received both her Master's degree and Ph.D. from the Department of Mathematics, IIT Roorkee, India. Her doctoral work focused on PDE-based image processing methods integrated with neural network frameworks. From 2023 to 2024, she conducted research in IMT Atlantique, France under the ANR-DREAM project, where she developed deep learning approaches to reconstruct and forecast satellite derived Chlorophyll-a concentration. Her research integrates mathematics, image processing, and environmental sciences to address key challenges in spatio-temporal modeling and prediction.
\end{IEEEbiographynophoto}
\begin{IEEEbiography}[{\includegraphics[width=1in,height=1.25in,clip,keepaspectratio]{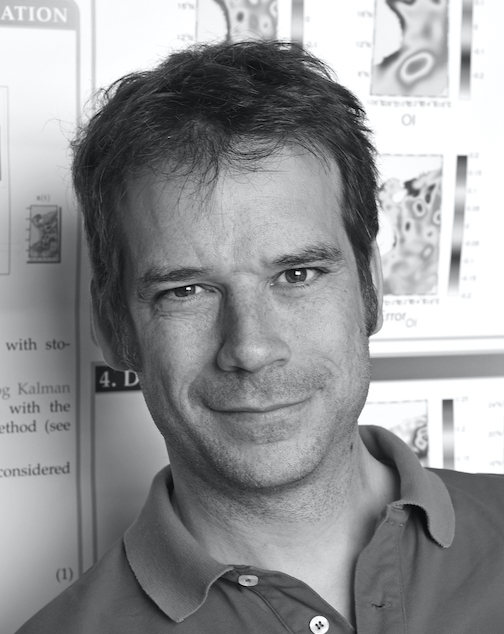}}]{Ronan Fablet}
 is Full Professor at IMT Atlantique and research scientist in AI for ocean science and is strongly involved in interdisciplinary research with oceanographers and marine ecology. He brings his expertise in theory-guided deep learning for the identification, forecasting and reconstruction of geophysical dynamics.
\end{IEEEbiography}

\begin{IEEEbiography}[{\includegraphics[width=1in,height=1.25in,clip,keepaspectratio]{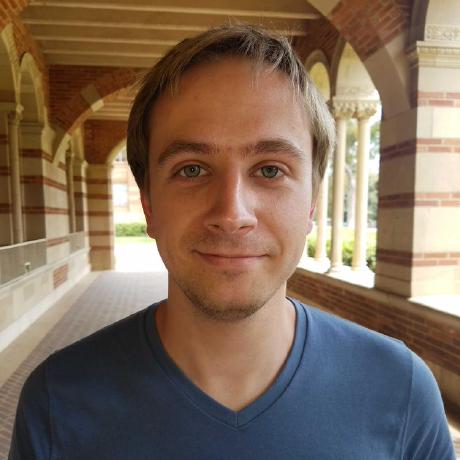}}]{Lucas Drumetz}
 is Ass. Professor at IMT Atlantique and research scientist at Lab-STICC, investigating physically consistent AI-based dynamical representations of multispectral image time series. His expertise covers machine learning for remote sensing applications (particularly satellite images and time series).\end{IEEEbiography}

\begin{IEEEbiography}[{\includegraphics[width=1in,height=1.25in,clip,keepaspectratio]{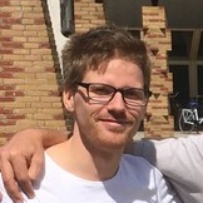}}]{Etienne Pauthenet}
is a doctor in physical oceanography. Over the last years, he has developed a strong expertise in applying machine learning techniques to the study of physical oceanography and biogeochemistry, drawing on both satellite and in situ observations to better understand ocean dynamics and processes. \end{IEEEbiography}

 \begin{IEEEbiography}[{\includegraphics[width=1in,height=1.25in,clip,keepaspectratio]{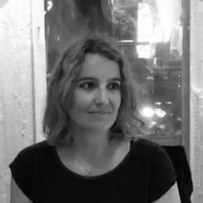}}]{Elodie Martinez}
is a senior researcher in physical and biogeochemistry oceanography. She has specific skills in satellite radiometric observations, investigating phytoplankton variability in regards to bottom-up process.  She is the leader of the ANR DREAM project\end{IEEEbiography}

\end{document}